\documentclass[sigconf]{aamas} 
\usepackage[ruled, vlined, linesnumbered]{algorithm2e}
\usepackage{amsmath}
\usepackage{dblfloatfix}
\usepackage{caption}
\usepackage{subcaption}
\usepackage{booktabs}
\usepackage{framed}
\usepackage{paralist}
\usepackage{booktabs}
\usepackage{multirow}
\usepackage[T1]{fontenc}
\usepackage{lmodern}

\usepackage{balance} 

\setcopyright{ifaamas}
\acmConference[AAMAS '23]{Proc.\@ of the 22nd International Conference
on Autonomous Agents and Multiagent Systems (AAMAS 2023)}{May 29 -- June 2, 2023}
{London, United Kingdom}{A.~Ricci, W.~Yeoh, N.~Agmon, B.~An (eds.)}
\copyrightyear{2023}
\acmYear{2023}
\acmDOI{}
\acmPrice{}
\acmISBN{}

\acmSubmissionID{541}

\title[HYDRA]{Learning to Operate in Open Worlds by \\ Adapting Planning Models}
\author{Wiktor Piotrowski}
\affiliation{
  \institution{Palo Alto Research Center}
  \city{Palo Alto}
  \country{USA}}
\email{wiktorpi@parc.com}
\author{Roni Stern}
\affiliation{
  \institution{Ben-Gurion University}
  \city{Beersheeba}
  \country{Israel}}
\email{wiktorpi@parc.com}
\author{Yoni Sher}
\affiliation{
  \institution{Palo Alto Research Center}
  \city{Palo Alto}
  \country{USA}}
\email{yoni.sher@parc.com}
\author{Jacob Le}
\affiliation{
  \institution{Palo Alto Research Center}
  \city{Palo Alto}
  \country{USA}}
\email{jale@parc.com}
\author{Matthew Klenk}
\affiliation{
  \institution{Toyota Research Institute}
  \city{Los Altos}
  \country{USA}}
\email{matt.klenk@tri.global}
\author{Johan deKleer}
\affiliation{
  \institution{Palo Alto Research Center}
  \city{Palo Alto}
  \country{USA}}
\email{dekleer@parc.com}
\author{Shiwali Mohan}
\affiliation{
  \institution{Palo Alto Research Center}
  \city{Palo Alto}
  \country{USA}}
\email{smohan@parc.com}

\vspace{-1cm}
\begin{abstract}
Planning agents are ill-equipped to act in novel situations in which their domain model no longer accurately represents the world. We introduce an approach for such agents operating in open worlds that detects the presence of novelties and effectively adapts their domain models and consequent action selection. It uses observations of action execution and measures their divergence from what is expected, according to the environment model, to infer existence of a novelty. Then, it revises the model through a heuristics-guided search over model changes. We report empirical evaluations on the CartPole problem, a standard Reinforcement Learning (RL) benchmark. The results show that our approach can deal with a class of novelties very quickly and in an interpretable fashion.
\end{abstract}

\newcommand{\BibTeX}{\rm B\kern-.05em{\sc i\kern-.025em b}\kern-.08em\TeX}
\newcommand{\tuple}[1]{\ensuremath{\langle #1 \rangle}}

\begin{document}


\pagestyle{fancy}
\fancyhead{}


\maketitle 

\setlength{\textfloatsep}{-01pt}
\vspace{-0.2cm}
\section{Introduction}
Artificial intelligence and machine learning research on sequential decision-making usually relies on the \emph{closed world} assumption. That is, all relevant characteristics of the environment are known ahead of deployment, during agent design time. 
For a decision-making agent that relies on automated planning techniques, 
knowledge about environmental characteristics is encoded explicitly as a domain model (description of actions, events, processes) that govern the agent's beliefs about the environment's dynamics. 
In an \emph{open world}, however, the characteristics of the environment often change while the agent is operational \cite{langley2020open, AI_for_openworlds_2022}. 
Such changes --- \emph{novelties} --- can cause a planning agent to fail catastrophically as its knowledge of the environment may become incomplete or incorrect. We explore how planning agents can robustly handle such novelties in an open world. 
Agents following our design use the planning domain model to also evaluate if observed outcomes diverge from what is expected in the plans it generated. If the divergence is significant, the novelty is inferred and accomodated through heuristics search.
This approach is applicable to planning agents implementing various levels of PDDL. Results in this paper are from a system implemented using PDDL+\cite{fox2006modelling} for CartPole \cite{brockman2016}, a classic control problem.

\vspace{-0.3cm}
\section{Approach}
\begin{figure}[t]
\vspace{-0.3cm}
    \centering
    \includegraphics[width=1\columnwidth]{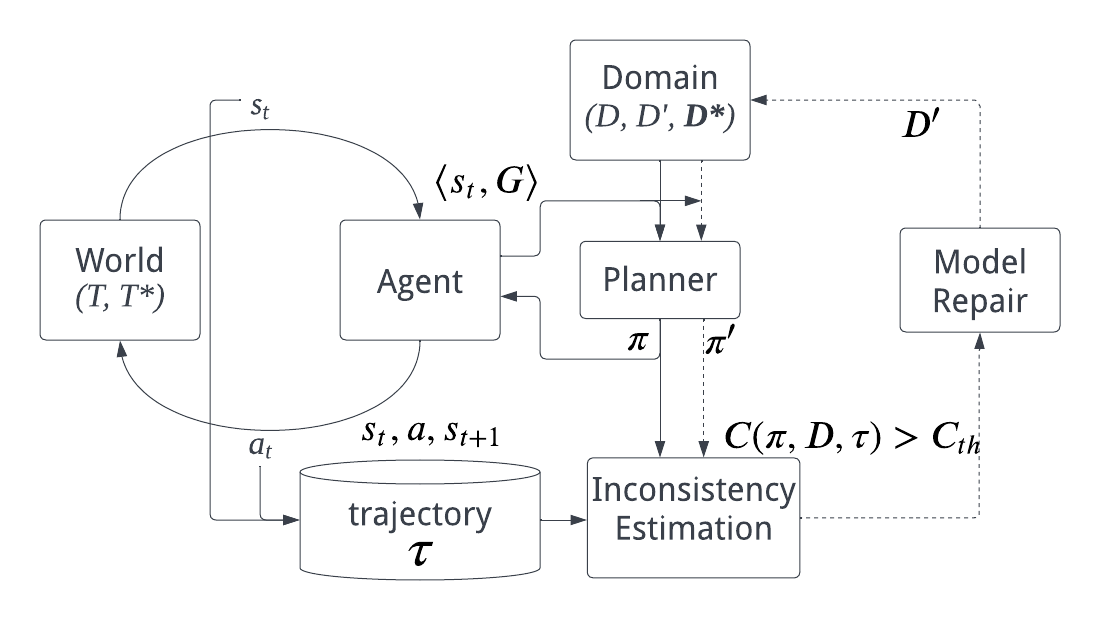}
    \vspace{-1cm}
    \caption{Diagram of novelty reasoning. Solid lines denote the planning process and dotted denote domain model revision.}
    \label{fig:novelty_reasoning}
\end{figure}
Figure \ref{fig:novelty_reasoning} shows the proposed agent design and the novelty reasoning process. 
The agent interacts with its environment in a sequence of episodes, where each episode is a set of actions taken by the agent to reach a terminal state. 
At some episode novelty is introduced and the environment changes, the agent is oblivious to the existence, timing, and nature of the introduced novelty. 

\begin{figure*}[hb]
    \centering
    \includegraphics[width=0.98\textwidth]{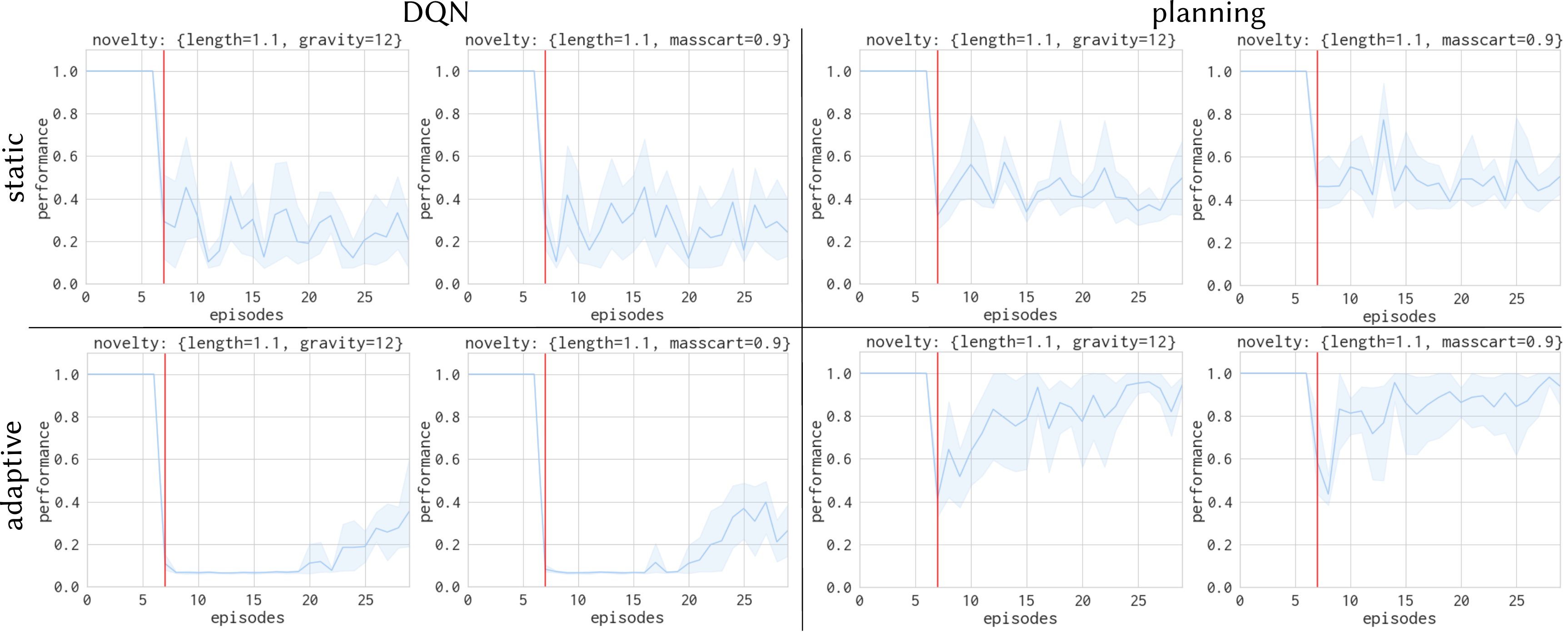}
    \vspace{-0.4cm}
    \caption{Graphs showing performance of DQN-static/adaptive and planning-static/repairing agents. Episodes are on the x-axis and reward on the y-axis. The results are averaged over $5$ trials. Red line indicates the episode $7$ when novelty was introduced.}
    \label{fig:combined_results}
\end{figure*}
At an episode's beginning, the agent accepts the current state $s_t$ and creates a corresponding planning problem ($s_t, G$) which is then paired with the domain model $D$. Then, it uses a planner to solve the problem to obtain plan $\pi$ and attempts to execute in the environment. During execution, it stores the observed trajectory $\tau$ as a list of $\tuple{s_t,a,s_{t+1}}$. At the episode's end, it computes an \emph{inconsistency score} for the current model $D$ by comparing the expected state trajectory with the observed execution trace, $\tau$.


Formally, let $S(\tau)$ be the sequence of states in observations and $S(\pi,D)$ be the expected sequence of states obtained by simulating the generated plan $\pi$ with the domain model $D$. 
Let $S(x)[i]$ denote the $i^{th}$ state in the state sequence $S(x)$. 
The inconsistency score is computed as
    $C(\pi, D, \tau) = \sum_{i} \gamma^i\cdot ||S(\tau)[i] - S(\pi,D)[i]||$
where $0<\gamma<1$ is a discount factor intended to limit the impact of sensing noise. 
If the inconsistency score exceeds a set threshold $C_{th}$, the agent infers that its domain model $D$ has become inconsistent with the novel environment characteristics. 
Then, it initiates the \emph{search-based model repair} process described in Algorithm \ref{alg:repair}
to adjust $D$ accordingly. 
Algorithm \ref{alg:repair} works by searching for a \emph{domain repair} $\varPhi$, which is a sequence of model modifications that, when applied to the agent's internal domain $D$, returns a domain $D'$ that is consistent with observations. To find such a repair, the algorithm accepts as input a set of basic \emph{Model Manipulation Operators} (MMOs), denoted $\{\varphi\} = \{\varphi_0, \varphi_1, ... , \varphi_n\}$. Each MMO $\varphi_i \in \{\varphi\}$ represents a possible change to the domain. A domain repair $\varPhi$ is a sequence of one or more basic MMO $\varphi_i \in \{\varphi\}$. An MMO example is to add an amount $\Delta\in\mathbb{R}$ to a numeric domain fluent. After this repair, the agent moves on to the next episode and uses the updated internal domain model $D'$ to solve the subsequent tasks. It may take a few repair steps to find a consistent domain model because a single trajectory may not provide enough information to find the correct repair. 
\vspace{-0.4cm}
\section{Results}
We evaluated our approach using a standard implementation of CartPole \cite{brockman2016}, where the task is to balance the pole in the upright position for $n=200$ steps by pushing the cart either left or right. The environment provides information on the velocities and positions of the cart and the pole (4-tuple). The domain's system dynamics are defined by several parameters: mass of the cart, mass of the pole, length of the pole, gravity, angle limit, cart limit, push force. 

\begin{algorithm}
\scriptsize
\SetKwInOut{Input}{Input}\SetKwInOut{Output}{Output}
\Input{$\{\varphi\}$: a set of basic MMOs; $D$: the original PDDL+ domain; $\pi$: plan generated using $D$; $\tau$: a trajectory; $C_{th}$: consistency threshold}
\Output{$\varPhi_{best}$, a domain repair for $D$}
OPEN$\gets\{\emptyset\}$; 
$C_{best}\gets\infty$;
$\varphi_{best}\gets\emptyset$\\
\While{$C_{best}\geq C_{th}$}{
    $\varPhi\gets$ pop from OPEN\\
    \ForEach{$\varphi_i\in\{\varphi\}$}{
        $\varPhi'\gets \varPhi \cup \varphi_i$ {\scriptsize \tcc*{Compose a domain repair}} 
        DoRepair($\varPhi'$, $D$)\\
        $C_{\varPhi'}\gets$ InconsistencyEstimator($\pi$, $D$, $\tau$)\\
        \If{$C_\varPhi\leq C_{best}$}{
            $C_{best}\gets C_{\varPhi'}$\\
            $\varPhi_{best}\gets\varPhi'$
        }
        Insert $\varPhi'$ to OPEN with key $f(\varPhi', C_{\varPhi'})$ \nllabel{alg:line:f}\\
        UndoRepair($\varPhi'$, $D$)
    }
}
\caption{PDDL+ model repair algorithm.}
\label{alg:repair}
\end{algorithm}

Figure \ref{fig:combined_results} summarizes the performance of various agents. We studied two novelties: changing the pole length to $1.1$ and the gravity to $12$; and changing the pole length to $1.1$ and the cart mass to $0.9$. 
As baselines, we implemented RL agents that use standard dynamic q-networks with experience replay\cite{mnih2013playing}. DQN-static employs the policy learned in non-novel settings in the novel environment while the dynamic version learns a new policy. 

The results show that the planning agents are, first, \emph{resilient}; the impact of novelty on their performance is not as drastic as on the DQN agents. 
It is because planning agents use models that are modular, composable and are written in a general way. 
In the novelty setting, a subset of model elements are still relevant. Second, our approach, the planning-adaptive agent learns \emph{quickly} and recovers optimal performance in around $20$ episodes. This observation supports our central thesis: model-space search enables quick adaptation in dynamic environments because it can localize the learning to specific parts of the explicit model and other parts are \emph{transferred}. In contrast, a DQN agent has to learn new network parameters afresh. Finally, the adaptations are \emph{interpretable}; they are expressed in the same language as the original model, enabling a model designer to inspect what the system has learned. Our method found the following example repairs for CartPole. Eaach element in the repair is a numeric domain fluent and the reported value is a change from its nominal value. \\
\small
\noindent\textbf{Repair 1:} \texttt{mass\_cart:0,length\_pole:0.3,mass\_pole:0,\\
 force\_mag:0,gravity:0,angle\_limit:0,x\_limit:0} \\
\textbf{Repair 2:} \texttt{mass\_cart:0,length\_pole:0,mass\_pole:0,force\_mag:0, \\
 gravity:1.0,angle\_limit:0,x\_limit:0}

 \subsection*{Acknowledgements}
The work presented in this paper was supported in part by the DARPA SAIL-ON program under award number HR001120C0040. The views, opinions and/or findings expressed are those of the authors’ and should not be interpreted as representing the official views or policies of the Department of Defense or the U.S. Government.

\bibliographystyle{ACM-Reference-Format} 
\bibliography{bibliography}


\end{document}